\documentclass[11pt]{article}
\usepackage{fullpage}
\usepackage{times}
\usepackage{url}
\usepackage{latexsym}
\usepackage{amsfonts}
\usepackage{multirow}
\usepackage{hyperref}
\hypersetup{
  colorlinks = true,
  citecolor = blue
}
\usepackage{booktabs}
\usepackage{mathtools}
\usepackage{enumitem}
\usepackage{color}
\usepackage{amsmath}
\usepackage{amssymb}
\usepackage{natbib}
\usepackage{xcolor}

\newcommand*\samethanks[1][\value{footnote}]{\footnotemark[#1]}

\def\R{\mathbb{R}}

\def\M{\mathcal{M}}

\newcommand{\norm}[1]{\left\|#1\right\|}

\def\bB{\mathbf B}

\def\bI{{\mathbf I}}

\def\bU{{\mathbf U}}
\def\bV{{\mathbf V}}

\def\bX{{\mathbf X}}
\def\bY{{\mathbf Y}}
\def\bZ{{\mathbf Z}}

\def\minop{\mathop{\rm min}\limits}

\begin{document}
\title{\LARGE \bf Learning Geometric Word Meta-Embeddings
}

\author{Pratik Jawanpuria\thanks{Microsoft India. Email: \{pratik.jawanpuria, ankunchu, bamdevm\}@microsoft.com.} \quad N T V Satya Dev\thanks{Email: tvsatyadev@gmail.com.}\quad Anoop Kunchukuttan\samethanks[1] \quad Bamdev Mishra\samethanks[1]}

\date{}

\maketitle

\begin{abstract}
 We propose a geometric framework for learning meta-embeddings of words from different embedding sources. Our framework transforms the embeddings into a common latent space, where, for example, simple averaging of different embeddings (of a given word) is more amenable. The proposed latent space arises from two particular geometric transformations - the orthogonal rotations and the Mahalanobis metric scaling. Empirical results on several word similarity and word analogy benchmarks illustrate the efficacy of the proposed framework. 
\end{abstract}

\section{Introduction}
Word embeddings have become an integral part of modern NLP. They capture semantic and syntactic similarities and are typically used as features in training NLP models for diverse tasks like named entity tagging, sentiment analysis, and classification, to name a few. Word embeddings are learnt in an unsupervised manner from a large text corpora and a number of pre-trained embeddings are readily available. The quality of the word embeddings, however, depends on various factors like the size and genre of training corpora as well as the training method used. This has led to ensemble approaches for creating meta-embeddings from different original embeddings \citep{yin16,bollegala18naacl,bollegala18iccl,oneilly2020meta}. Meta-embeddings are appealing because: (a) they can improve quality of embeddings on account of noise cancellation and diversity of data sources and algorithms, (b) no need to retrain the model, (c) the original corpus may not be available, and (d) may increase vocabulary coverage. 

Various approaches have been proposed to learn meta-embeddings and can be broadly classified into two categories: (a) simple linear methods like averaging or concatenation, or a low-dimensional projection via singular value projection \citep{yin16,bollegala18naacl} and (b) non-linear methods that aim to learn meta-embeddings as shared representation using auto-encoding or transformation between common representation and each embedding set \citep{muromagi-etal-2017-linear,bollegala2017think,bollegala18iccl,oneilly2020meta}. 



In this work, we focus on simple linear methods such as averaging and concatenation for computing meta-embeddings, which are very easy to implement and have shown highly competitive performance \citep{yin16,bollegala18naacl}. Due to the nature of the underlying embedding generation algorithms \citep{mikolov13bnips,pennington14}, correspondences between dimensions, e.g., of two embeddings $x\in\R^d$ and $z\in\R^d$ of the same word, are usually not known. Hence, averaging may be detrimental in cases where the dimensions are negatively correlated. Consider the scenario where $z\coloneqq-x$. Here, simple averaging of $x$ and $z$ would result in the zero vector. Similarly, when $z$ is a (dimension-wise) permutation of $x$, simple averaging would result in a sub-optimal meta-embedding vector than performing averaging of \textit{aligned} embeddings. Therefore, we propose to align the embeddings (of a given word) as an important first step towards generating meta-embeddings. 

To this end, we develop a geometric framework for learning meta-embeddings, by aligning different embeddings in a common latent space, where the dimensions of different embeddings (of a given word) are in coherence. Mathematically, we perform different orthogonal transformations of the source embeddings to learn a latent space along with a Mahalanobis metric that scales different features appropriately. The meta-embeddings are, subsequently, learned in the latent space, e.g., using averaging or concatenation. Empirical results on the word similarity and the word analogy tasks show that the proposed geometrically aligned meta-embeddings outperform strong baselines such as the plain averaging and the plain concatenation models.

\section{Proposed Geometric Modeling}

Consider two (monolingual) embeddings $x_i\in\R^d$ and $z_i\in\R^d$ of a given word $i$ in a $d$-dimensional space. As discussed earlier, embeddings generated from different algorithms \citep{mikolov13bnips,pennington14} may express different characteristics (of the same word). Hence, the goal of learning a meta-embedding $w_i$ (corresponding to word $i$) is to generate a representation that inherits the properties of the different source embeddings (e.g., $x_i$ and $z_i$).

Our framework imposes orthogonal transformations on the given source embeddings to enable alignment. In this latent space, we additionally induce the Mahalanobis metric to incorporate the feature correlation information \citep{mjaw19a}. The Mahalanobis similarity generalizes the cosine similarity measure, which is commonly used for evaluating the relatedness between word embeddings. The combination of the orthogonal transformation and Mahalanobis metric learning allows to capture any \textit{affine} relationship between different available source embeddings of a given word \citep{bonnabel09a,mishra14}.



Overall, we formulate the problem of learning geometric transformations -- the orthogonal rotations and the metric scaling -- via a binary classification problem. The meta-embeddings are subsequently computed using these transformations. The following sections formalize the proposed latent space and meta-embedding models.



\subsection{Learning the Latent Space}
In this section, we learn the latent space using geometric transformations.

Let $\bU\in \M^d$ and $\bV \in \M^d$ be orthogonal transformations for embeddings $x_i$ and $z_i$, respectively, for all words $i$. Here $\M^d$ represents the set of $d\times d$ orthogonal matrices. 
The aligned embeddings in the latent space corresponding to $x_i$ and $z_i$ can then be expressed as $\bU x_i$ and $\bV z_i$, respectively. We next induce the Mahalanobis metric $\bB$ in this (aligned) latent space, where $\bB$ is a $d\times d$ symmetric positive-definite matrix. In this latent space, the similarity between the two embeddings $x_i$ and $z_i$ is given by the following expression: $(\bU x_i)^\top \bB (\bV z_i)$. An equivalent interpretation is that the expression $(\bU x_i)^\top \bB (\bV z_i)$ boils down to the standard scalar product (cosine similarity) between $\bB^\frac{1}{2}\bU x_i$ and $\bB^\frac{1}{2}\bV z_i$, where $\bB^\frac{1}{2}$ denotes the matrix square root of the symmetric positive definite matrix $\bB$.



The orthogonal transformations as well as the Mahalanobis metric are learned via the following binary classification problem: pairs of word embeddings $\{x_i,z_i\}$ of the same word $i$ belong to the positive class while pairs $\{x_i,z_j\}$ belong to the negative class (for $i\neq j$). We consider the similarity between the two embeddings in the latent space as the decision function of the proposed binary classification problem. Let $\bX = [x_1,\ldots,x_n]\in\R^{d\times n}$ and $\bZ = [z_1,\ldots,z_n]\in\R^{d\times n}$ be the word embedding matrices for $n$ words, where the columns correspond to different words. In addition, let $\bY$ denote the label matrix, where $\bY_{ii}=1$ for $i=1,\ldots,n$ and $\bY_{ij}=0$ for $i\neq j$. The proposed optimization problem employs the simple to optimize square loss function: 
\begin{equation}\label{eqn:geommavg}
  \minop_{\substack{\bU,\bV\in\M^d,\\\bB\succ0}} \norm{\bX^\top\bU^\top\bB\bV\bZ - \bY}^2 + C\norm{\bB}^2,
\end{equation}
where $\|\cdot\|$ is the Frobenius norm and $C>0$ is the regularization parameter.




\subsection{Averaging and Concatenation in Latent Space}

Meta-embeddings constructed by averaging or concatenating the given word embeddings have been shown to obtain highly competitive performance \citep{yin16,bollegala18naacl}. Hence, we propose to learn meta-embeddings as averaging or concatenation in the learned latent space.

\subsubsection*{Geometry-Aware Averaging}
The meta-embedding $w_i$ of a word $i$ is generated as an average of the (aligned) word embeddings in the latent space. The latent space representation of $x_i$, as a function of orthogonal transformation $\bU$ and metric $\bB$, is $\bB^\frac{1}{2}\bU x_i$ \citep{mjaw19a}.  
Hence, we obtain $w_i = \texttt{average}(\bB^\frac{1}{2}\bU x_i,\bB^\frac{1}{2}\bV z_i)=(\bB^\frac{1}{2}\bU x_i+\bB^\frac{1}{2}\bV z_i)/2$. 

It should be noted that the proposed geometry-aware averaging approach generalizes \citep{bollegala18naacl}, which is now a particular case in our framework by choosing $\bU$, $\bV$, and $\bB$ as identity matrices. 
Our framework easily generalizes to the case of more than two source embeddings, by learning different source-embedding specific orthogonal transformations and a common Mahalanobis metric.


\subsubsection*{Geometry-Aware Concatenation}
We next propose to concatenate the aligned embeddings in the learned latent space. For a given word $i$, with $x_i$ and $z_i$ as different source embeddings, the meta-embeddings $w_i$ learned by the proposed geometry-aware concatenation model is $w_i=\texttt{concatenation}(\bB^\frac{1}{2}\bU x_i,\bB^\frac{1}{2}\bV z_i)=[(\bB^\frac{1}{2}\bU x_i)^\top,(\bB^\frac{1}{2}\bV z_i)^\top]^\top$. It can be easily observed that the plain concatenation \citep{yin16} is a special case of the proposed geometry-aware concatenation (by setting $\bB=\bU=\bV=\bI$, where $\bI$ is a $d$-dimensional identity matrix).

\subsection{Optimization}
The proposed optimization problem (\ref{eqn:geommavg}) employs square loss function and $\ell_2$-norm regularization, both of which are well-studied in literature. In addition, the proposed problem involves optimization over smooth constraint sets such as the set of symmetric positive definite matrices and the set of orthogonal matrices. Such sets have well-known Riemannian manifold structure \citep{Lee03a} that allows to propose computationally efficient iterative optimization algorithms. We employ the popular Riemannian optimization framework \citep{absil08a} to solve (\ref{eqn:geommavg}). Recently, \citet{mjaw19a} have studied a similar optimization problem in the context of learning cross-lingual word embeddings. 

Our implementation is done using the Pymanopt toolbox \citep{townsend16a}, which is a publicly available Python toolbox for Riemannian optimization algorithms. In particular, we use the conjugate gradient algorithm of Pymanopt. For, this we need only supply the objective function of (\ref{eqn:geommavg}). This can be done efficiently as the numerical cost of computing the objective function is $O(nd^2)$. The overall computational cost of our implementation scales linearly with the number of words in the vocabulary sets.

\begin{table*}[t]
  \setlength{\tabcolsep}{3.5pt}
  
  \begin{center}
    \begin{tabular}{l|l|ccccccc|cccc}
      \toprule
      & Model & RG & MC & WS & MTurk & RW & SL & Avg.(WS) & MSR & GL & SemEvaL & Avg.(WA) \\
      \midrule
      \parbox[t]{2mm}{\multirow{3}{*}{\rotatebox[origin=c]{90}{indv.}}} &  CBOW  &  $76.1$  &  $80.0$  &  $77.2$  &  $68.4$  &  $53.4$  &  $44.2$  &  $66.5$  &  $71.7$  &  $55.4$  &  $20.4$  &  $49.2$\\
      &  GloVe  &  $82.9$  &  $84.0$  &  $79.6$  &  $70.0$  &  $48.7$  &  $45.3$  &  $68.4$  &  $69.3$  &  $75.2$  &  $18.6$  &  $54.4$\\
      &  fastText  &  $83.8$  &  $82.5$  &  $83.5$  &  $73.3$  &  $58.0$  &  $46.4$  &  $71.2$  &  $78.7$  &  $71.0$  &  $22.5$  &  $57.4$\\
      \midrule
      \parbox[t]{2mm}{\multirow{4}{*}{\rotatebox[origin=c]{90}{Gl.+ CB.}}}  &  CONC  &  $81.1$  &  $84.6$  &  $\mathbf{81.4}$  &  $71.9$  &  $54.6$  &  $46.0$  &  $69.9$  &  $76.6$  &  $69.9$  &  $\mathbf{20.1}$  &  $55.5$\\
      &  AVG  &  $81.5$  &  $83.7$  &  $79.4$  &  $\mathbf{72.1}$  &  $52.9$  &  $46.2$  &  $69.3$  &  $73.7$  &  $66.9$  &  $19.7$  &  $53.4$\\
      &  Geo-CONC  &  $\mathbf{86.0}$  & $\mathbf{85.0}$  & $81.2$  & $70.5$  & $55.6$  & $\mathbf{48.2}$  & $\mathbf{71.1}$  & $\mathbf{78.1}$  & ${73.3}$  & $19.9$  & $\mathbf{57.1}$\\
      &  Geo-AVG  &  $85.8$  &  $83.5$  &  $81.2$  &  $69.1$  &  $\mathbf{55.7}$  &  $\mathbf{48.2}$  &  $70.6$  &  $77.3$  &  $72.3$  &  $19.5$  &  $56.3$\\

      \midrule
      \parbox[t]{2mm}{\multirow{4}{*}{\rotatebox[origin=c]{90}{Gl.+ fa.}}} &  CONC  &  $\mathbf{83.8}$  &  $82.5$  &  $83.4$  &  $73.3$  &  $57.9$  &  $46.4$  &  $71.2$  &  $79.8$  &  $71.7$  &  $22.5$  &  $58.0$\\
      &  AVG  &  $83.4$  &  $82.1$  &  $\mathbf{83.5}$  &  $73.3$  &  $\mathbf{58.0}$  &  $46.5$  &  $71.1$  &  $79.7$  &  $71.7$  &  $22.4$  &  $57.9$\\
      &  Geo-CONC  &  $83.7$  &  $\mathbf{84.0}$  &  $82.6$  &  $\mathbf{74.6}$  &  $55.1$  &  $\mathbf{48.4}$  &  $\mathbf{71.4}$  &  $\mathbf{80.4}$  &  $\mathbf{79.3}$  &  $21.5$  &  $\mathbf{60.4}$\\
      &  Geo-AVG  &  $83.6$  &  $82.0$  &  $82.7$  &  $74.3$  &  $57.0$  &  $\mathbf{48.4}$  &  $71.3$  &  $79.1$  &  $71.1$  &  $\mathbf{23.1}$  &  $57.8$\\

      \midrule
      \parbox[t]{2mm}{\multirow{4}{*}{\rotatebox[origin=c]{90}{CB.+ fa.}}}  &  CONC  &  $83.8$  &  $82.5$  &  $\mathbf{83.5}$  &  $73.6$  &  $\mathbf{59.9}$  &  $46.4$  &  $71.6$  &  $79.9$  &  $75.8$  &  $\mathbf{22.5}$  &  $59.4$\\
      &  AVG  &  $83.7$  &  $82.5$  &  $83.4$  &  $\mathbf{73.7}$  &  $59.8$  &  $46.4$  &  $71.6$  &  $79.9$  &  $75.8$  &  $\mathbf{22.5}$  &  $59.4$\\
     
     
      &  Geo-CONC   &  $85.3$  &  $84.3$  &  $82.9$  &  $73.6$  &  $59.7$  &  $\mathbf{47.4}$  &  $72.2$  &  $\mathbf{80.1}$  &  $\mathbf{76.9}$  &  $22.1$  &  $\mathbf{59.7}$\\
      &  Geo-AVG  &  $\mathbf{85.5}$  &  $\mathbf{84.6}$  &  $82.9$  &  $73.6$  &  $59.7$  &  $\mathbf{47.4}$  &  $\mathbf{72.3}$  &  $79.9$  &  $\mathbf{76.9}$  &  $22.0$  &  $59.6$\\
      \bottomrule
    \end{tabular}
    \caption{Generalization performance of the meta-embedding algorithms on the Word Similarity (WS) and Word Analogy (WA) tasks. The columns `Avg.(WS)' and `Avg.(WA)' correspond to the average performance on the WS and the WA tasks, respectively. The rows marked `indv.' correspond to the performance of individual source embeddings CBOW, GloVe, and fastText. The rows marked `Gl.+CB.' correspond to the performance of meta-embedding algorithms with GloVe and CBOW embeddings as input. Similarly, `Gl.+fa.' corresponds to GloVe and fastText embeddings and `CB.+fa.' implies CBOW and fastText embeddings. A meta-embedding result is highlighted if it obtains the best result on a dataset when compared with the corresponding source embeddings as well as other meta-embedding algorithms employing the same source embeddings. We observe that the best overall performance on both the tasks, word similarity and word analogy, is obtained by the proposed geometry-aware models for every pair of input source embeddings.}\label{table:geometa}
  \end{center}
\end{table*}

\section{Experiments}
In this section, we evaluate the performance of the proposed meta-embedding models. 

\subsection{Evaluation Tasks and Datasets} 

We consider the following standard evaluation tasks \citep{yin16,bollegala18naacl}:
\begin{itemize}
    \item \textbf{Word similarity}: in this task, we compare the human-annotated similarity scores between pairs of words with the corresponding cosine similarity computed via the constructed meta-embeddings. We report results on the following benchmark datasets: {\bf RG} \citep{rubenstein}, {\bf MC} \citep{miller}, {\bf WS} \citep{finkelstein}, {\bf MTurk} \citep{halawi}, {\bf RW} \citep{luong}, and {\bf SL} \citep{hill}. Following previous works \citep{yin16,bollegala18naacl,oneilly2020meta}, we report the Spearman correlation score (higher is better) between the cosine similarity (computed via meta-embeddings) and the human scores. 
    \item \textbf{Word analogy}: in this task, the aim is to answer questions which have the form ``\textit{A is to B as C is to ?}'' \citep{mikolov13bnips}. After generating the meta-embeddings $a$, $b$, and $c$ (corresponding to terms \textit{A}, \textit{B}, and \textit{C}, respectively), the answer is chosen to be the term whose meta-embedding has the maximum cosine similarity with $(b-a+c)$ \citep{mikolov13bnips}.
    The benchmark datasets include {\bf MSR} \citep{msrgao}, {\bf GL} \citep{mikolov13bnips}, and {\bf SemEval} \citep{davidsemeval}. Following previous works  \citep{yin16,bollegala18naacl,oneilly2020meta}, we report the percentage of correct answers for MSR and GL datasets, and the Spearman correlation score for SemEval. In both cases, a higher score implies better performance. 
\end{itemize}
We learn the meta-embeddings from the following publicly available $300$-dimensional pre-trained word embeddings for English. 
\begin{itemize}
    \item \textbf{CBOW} \citep{mikolov13bnips}: has $929\,023$ word embeddings trained on Google News.
    \item \textbf{GloVe} \citep{pennington14}: has $1\,917\,494$ word embeddings trained on $42$B tokens of web data from the common crawl.
    \item \textbf{fastText} \citep{bojanowski17}: has $2\,000
    \,000$ word embeddings trained on common crawl.
\end{itemize}
The meta-embeddings are learned on the common set of words from different pairs of the source embeddings. The number of common words between various source embeddings pairs are as follows: $154\,077$ (GloVe $\cap$ CBOW), $552\,168$ (GloVe $\cap $ fastText), and $641\,885$ (CBOW $\cap $ fastText).

\subsection{Results and Discussion} 
The performance of our geometry-aware averaging and concatenation models, henceforth termed as Geo-AVG and Geo-CONC, respectively, are reported in  Table~\ref{table:geometa}. 
We also report  the performance of meta-embeddings models AVG \citep{bollegala18naacl} and CONC \citep{yin16}, which perform plain averaging and concatenation, respectively. 
In addition, we report the performance of individual source embeddings (CBOW, GloVe, and fastText), serving as a benchmark which the meta-embeddings algorithms should ideally surpass in order to justify their usage. 

We observe that the proposed geometry-aware models, Geo-AVG and Geo-CONC, outperform the individual source embeddings in all the datasets. The proposed models also easily surpass the AVG and CONC models in both the word similarity and the word analogy tasks. This shows that the alignment of word embedding spaces with orthogonal rotations and the Mahalanobis metric improves the overall quality of the meta-embeddings. 



\section{Conclusion}
We propose a geometric framework for learning meta-embeddings of words from various sources of word embeddings. Our framework aligns the embeddings in a common latent space. The importance of learning the latent space is shown in several benchmark datasets, where the proposed algorithms (Geo-AVG and Geo-CONC) outperforms the plain averaging and the plain concatenation models. The proposed framework can be extended to generating sentence meta-embeddings, which remains a future research direction. 



\bibliographystyle{acl_natbib}
\bibliography{geometric_meta_embeddings}

\end{document}